\documentclass[pdflatex,sn-mathphys-ay]{sn-jnl}

\usepackage{graphicx}
\usepackage{amsmath,amssymb,amsfonts}
\usepackage{amsthm}
\usepackage[title]{appendix}
\usepackage{xcolor}
\usepackage{booktabs}
\usepackage{multirow}
\usepackage{url}

\theoremstyle{thmstyleone}

\begin{document}

\title[When Does Reallocating a Sampling Budget Pay?]{When Does Reallocating a Sampling Budget Pay? Instance-wise Allocation for Neural Routing Solvers}

\author*{\fnm{Jinhyung} \sur{Bae}}\email{kevinbae2006@hufs.ac.kr}

\affil*{\orgdiv{Department of Industrial and Management Engineering}, \orgname{Hankuk University of Foreign Studies}, \orgaddress{\city{Yongin}, \country{Republic of Korea}}}

\abstract{Neural routing solvers sample many solutions per instance and keep the best, with the same number of samples for every instance. We measure how much a non-uniform split of the same total budget is worth and which property of a workload decides it. Best-of-$k$ curves estimated by averaging random orderings of stored samples overstate the gain, and deciding on one half of the samples while evaluating on the other understates it; we use exact best-of-$k$ curves, a U-statistic over all orderings. For three pretrained solvers (POMO, the Attention Model and Sym-NCO) on the traveling salesman problem, reallocation reduces the total optimality gap by at most 2.5\% on the training distribution, so uniform allocation is close to optimal there. On workloads that mix in out-of-distribution instances it reduces the gap by 11.8--13.3\% (1.1--1.8\% of tour length) for the two solvers that degrade under the shift, and almost nothing for the one that does not. Across two heterogeneity mechanisms, three solvers, two confirmation collections and budgets of 25 to 200 samples per instance, the gain is predicted by the spread of difficulty within a workload, not its average. Allocating on a power-law curve fitted to twenty charged probe samples per instance recovers 88--93\% of the uncharged optimum; successive halving does worse. A 2-opt local search, however, removes most of the gap at negligible cost, after which reallocation adds a few hundredths of a percent of tour length.}


\maketitle

\section{Introduction}\label{sec:intro}

Constructive neural solvers for routing problems do not return one tour. They decode many and report the best. The Attention Model samples 1,280 tours per instance \citep{kool2019attention}; POMO starts a greedy rollout from every node under eight symmetry augmentations, 800 trajectories for a 100-node instance \citep{kwon2020pomo}; later methods spend their search effort in a fixed number of iterations per instance \citep{hottung2022eas,luo2023lehd} or a fixed sampling schedule per benchmark \citep{sun2023difusco}. In each case the amount of test-time computation is a single global setting, identical for every instance, and benchmark suites fix a time budget in advance per dataset or as a function of instance size \citep{thyssens2023arena}. This is a sensible convention for comparing methods: every method sees the same budget on every instance.

It is also an allocation decision taken by default. Instances do not respond equally to extra samples. For some, the best-of-$k$ curve is flat after a handful of samples; for others it keeps falling. If the marginal value of a sample differs across instances, a uniform split of a fixed total budget leaves something on the table. How much it leaves, and on which workloads, has to our knowledge not been measured for neural combinatorial optimization (NCO) solvers. The analogous question has been studied for language models, where adaptive best-of-$N$ allocation across prompts improves on uniform allocation at the same total budget \citep{damani2025hard,raman2025adabon}. Neural routing solvers differ in two ways that matter here: the objective is a continuous tour length rather than a verifier-checked success, and the solvers degrade in a characteristic way on instances unlike their training distribution \citep{liu2023howgood}.

This paper asks two questions. How much does instance-wise reallocation of a fixed sampling budget buy for NCO solvers? And which observable property of a workload predicts whether it pays? Answering the first turned out to depend on how the gain is measured, so the paper also makes a methodological contribution, in the tradition of this journal's concern with how heuristics are tested \citep{hooker1995testing}.

\paragraph{Findings.}
\begin{enumerate}
\item \textbf{Measurement.} The obvious estimator, which builds each instance's best-of-$k$ curve by averaging random orderings of its stored samples and then optimizes the allocation on those curves, overstates the gain by 0.7--1.5 points, because the optimizer exploits the Monte Carlo noise. The obvious correction, deciding on half the samples and evaluating on the other half, understates it, because half-pool curves are not full-pool curves. Exact curves avoid both problems: on a deterministic pool they are the true curves, which exposes the size of both biases, and under stochastic sampling their residual bias is small (Section~\ref{sec:measure}).
\item \textbf{Size of the gain.} On the training distribution, reallocation reduces the total gap by 1.1--1.6\% on a fixed solution pool and 0.7--2.5\% under stochastic sampling: uniform allocation is close to optimal. On a workload that mixes uniform and clustered instances it buys 13.3\% (Attention Model) and 11.8\% (Sym-NCO), and 0.5\% for POMO, whose quality barely degrades on the clustered instances (Section~\ref{sec:howmuch}).
\item \textbf{What predicts it.} Within a workload, the spread of instance difficulty predicts the gain; its average does not. This holds under two different heterogeneity mechanisms, three solvers and two confirmation collections. The average difficulty appears to matter but changes sign between mechanisms; we argue that it is more consistent with an effect of instance scale (Section~\ref{sec:predict}).
\item \textbf{A deployable rule.} A probe of twenty samples per instance, charged against the same budget, recovers 62--66\% of the uncharged optimum with a single probe statistic, and 88--93\% when a power-law curve is fitted to each probe and the allocation is refined in a second stage; on a solver the workload does not disturb both lose, the fitted rule by 1.35 points, more than that solver's whole oracle headroom of 0.51 (Section~\ref{sec:policy}).
\item \textbf{Local search.} A 2-opt step, at about 3 ms per tour, lowers the mean gap on the regenerated mixed workload from 15--28\% to 2.3--4.7\%, roughly ten times more than reallocation, and after it reallocation no longer helps (Section~\ref{sec:ls}). Reallocation matters only for pipelines that end at the sampled solutions.
\end{enumerate}

\paragraph{What we do not claim.} We do not claim causation: workloads are constructed, and instances are not randomly assigned to difficulty. We study one problem class, the traveling salesman problem, and do not claim that the results transfer to others. We do not claim that adaptive allocation, the exact best-of-$k$ estimator or probe-then-fit allocation is new: each is established in stratified sampling, heuristic search, hyperparameter optimization or language-model inference (Section~\ref{sec:related}). Our claim is the measurement for neural routing solvers, the identification of the workload property that governs it, the demonstration of opposite estimator biases on heuristic benchmarks, and a charged rule that works on this objective.

\paragraph{Relation to the author's preprints.} This paper supersedes two earlier preprints \citep{bae2026luck,bae2026spread}, re-analyses their stored data with the exact estimator, and corrects one of their conclusions. Appendix~\ref{app:preprints} lists what changed.

\section{Related work}\label{sec:related}

\paragraph{Test-time budgets in neural routing.} Sampling and augmentation \citep{kool2019attention,kwon2020pomo,kim2022symnco}, active search \citep{hottung2022eas}, simulation-guided beam search \citep{choo2022sgbs}, diffusion sampling \citep{sun2023difusco}, iterative reconstruction \citep{luo2023lehd} and memory-based online adaptation \citep{chalumeau2024memento} all improve how a budget is spent \emph{within} an instance. Benchmarking work evaluates solvers under fixed time budgets and anytime curves \citep{thyssens2023arena}, and guidelines for testing machine-learning methods on routing problems stress that instance structure drives difficulty \citep{accorsi2022guidelines}. In all of these the per-instance budget is fixed in advance and identical across instances of a given size. Apart from the preprints this paper supersedes, we found no study that reallocates a fixed total budget \emph{across} instances for neural routing solvers.

\paragraph{Extreme-value objectives in heuristic search.} Spending trials to obtain the best possible value is the subject of the max $k$-armed bandit, which allocates restarts among stochastic heuristics on a single instance so as to maximize the single best value found \citep{cicirello2005maxk,streeter2006maxk,streeter2006threshold}, and of its modern form, extreme bandits \citep{carpentier2014extreme,baudry2022extreme}. Multi-start strategies allocate computation among parallel local-search runs by their observed progress \citep{gyorgy2011multistart}, and extreme-value models of the best of many heuristic runs go back to \citet{golden1979interval}; \citet{ovacik2000integrating} combined such estimates with local search in this journal. These works share our best-of-$k$ objective but optimize one maximum over arms or runs on one instance, not a sum of per-instance minima over many instances.

\paragraph{Allocation across instances and for aggregate objectives.} Restart schedules have been learned for whole ensembles of instances \citep{streeter2007restart}, and solver portfolios allocate time among algorithms across sequences of instances \citep{gomes2001portfolios,gagliolo2006portfolios,kadioglu2011selection}; their objectives are time to solution or the number of instances solved, and the decision is a shared schedule rather than a per-instance sample count. The closest classical analogue of our setting is stratified sampling, where samples are spread across strata to estimate a weighted sum and the gain of optimal (Neyman) allocation over proportional allocation grows with how much the strata differ \citep{neyman1934representative,carpentier2011stratified}; our spread result plays the same role for minimum order statistics instead of variances. Optimal computing budget allocation \citep{chen2000ocba,chen2011ocba}, racing \citep{lopezibanez2016irace} and successive halving \citep{karnin2013sh,jamieson2016nonstochastic,li2018hyperband} spend a budget across alternatives or instances in order to identify the best one, which is why the latter performs poorly on our sum objective (Section~\ref{sec:policy}).

\paragraph{Adaptive test-time compute for language models.} Uneven allocation of a fixed sampling budget across inputs is well established for language models. Repeated sampling improves coverage roughly log-linearly in the number of samples \citep{brown2024monkeys}, and the aggregate power law arises from the distribution of per-problem difficulty \citep{schaeffer2025monkeys}, which parallels our finding that the spread of difficulty governs the gain. Methods predict per-input marginal benefit with a learned model \citep{damani2025hard}, set compute by estimated difficulty \citep{snell2025scaling}, spend a small exploration budget per prompt and distribute the rest \citep{raman2025adabon,nguyen2026uncertainty}, cast the problem as a bandit \citep{zuo2026strategic}, stop sampling early per query \citep{aggarwal2023adaptive,li2024escape}, or extrapolate pass@$k$ from few samples \citep{kazdan2025passk}. Fitting a parametric curve to a short prefix and allocating on the extrapolation is also standard for learning curves in hyperparameter optimization \citep{domhan2015extrapolation}. In the author's companion work on language models \citep{bae2026giveup}, the objective is binary and the value of one more attempt peaks at intermediate difficulty, so the average difficulty of a workload matters; the objective here is continuous, and Section~\ref{sec:discussion} returns to the difference.

\paragraph{Estimating best-of-$k$ curves and testing heuristics.} The exact expected best of a random $k$-subset of stored samples (Eq.~\eqref{eq:ustat}) is a U-statistic \citep{hoeffding1948ustat} with a textbook closed form \citep{david2003order}. In machine-learning evaluation it replaced the biased with-replacement estimate of expected best performance against budget \citep{dodge2019show,tang2020showing}, and in its 0/1 form it is the standard pass@$k$ estimator \citep{chen2021codex}. Optimizing a decision on noisy estimates and reporting its estimated value is biased upward, the optimizer's curse \citep{smith2006curse} and the optimistic bias of sample-average optima \citep{mak1999bounding}, while evaluating on held-out data corrects it but can underestimate the true value \citep{ito2018unbiased}. Section~\ref{sec:measure} shows both biases in estimating reallocation gains for heuristics and uses the existing exact estimator, in line with long-standing guidance on designing and reporting computational experiments with heuristics \citep{barr1995designing,hooker1995testing,johnson2002guide,bartzbeielstein2020benchmarking}. Our confirmation collections were pre-registered \citep{nosek2018prereg}; Appendix~\ref{app:prereg} reports how the registered verdicts change under the exact estimator.

\paragraph{What is new here.} Neither adaptive allocation nor probe-then-fit allocation is new. What this paper adds is (i) the measurement of the value of instance-wise reallocation for neural routing solvers, whose objective is a continuous best-of-$k$ tour length and whose convention is uniform per-instance budgets; (ii) evidence across mechanisms, solvers, collections and budgets that the value depends on the spread of difficulty within a workload and not on its average; (iii) a demonstration on heuristic benchmarks that the two usual ways of estimating the gain err in opposite directions; and (iv) a probe-and-fit rule, charged for its own probe, that recovers about 90\% of the attainable gain where a best-arm method such as successive halving does not.

\section{Problem and measurement}\label{sec:measure}

\subsection{The allocation problem}

A workload has $N$ instances and a total budget of $S = uN$ samples, where $u$ is the uniform per-instance budget. Let $f_i(k)$ be the expected gap to a reference tour of the best of $k$ samples for instance $i$, in percent. The allocation problem is
\begin{equation}
\min_{k_1,\dots,k_N}\ \sum_{i=1}^N f_i(k_i) \quad \text{s.t.}\quad \sum_{i=1}^N k_i = S,\quad 1 \le k_i \le c,\quad k_i\in\mathbb{Z},
\label{eq:alloc}
\end{equation}
where $c=400$ is a per-instance cap used throughout. Each $f_i$ is non-increasing and convex in $k$ (for the exact curve of Eq.~\eqref{eq:ustat}, the marginal reduction $\hat f(k)-\hat f(k+1)$ equals $1/(k+1)$ times the expected gap between the two smallest values of a random $(k+1)$-subset; the resulting curves are convex for all 840 instances we checked, and our greedy implementation additionally enforces non-increasing marginals), being the expectation of a minimum order statistic, so the problem is separable and the greedy rule, which gives each next sample to the instance with the largest marginal reduction, is optimal. We report the \emph{headroom}
\begin{equation}
d = 100\,\frac{\sum_i f_i(u) - \sum_i f_i(k_i^\ast)}{\sum_i f_i(u)}\,,
\label{eq:headroom}
\end{equation}
the percentage reduction in total gap that the optimal allocation $k^\ast$ achieves over the uniform one. It is a share of the optimality gap, not of the tour length: a headroom of 10\% on a workload whose uniform gap is 5\% shortens the tours by about half a percent. If all $f_i$ are identical, $d=0$ by convexity. Some dependence of $d$ on how much instances differ therefore follows from the setup. What does not follow is which observable property of a workload tracks it, whether its average difficulty carries information once spread is held fixed, and how large $d$ is in practice.

\subsection{Estimating the curves}

The curves $f_i$ are unknown and must be estimated from stored samples. Suppose $m$ costs are stored for an instance and sorted, $x_{(1)}\le\dots\le x_{(m)}$. Three estimators are natural.

\emph{Monte Carlo curves.} Draw $R$ random orderings of the stored costs, take the running minimum of each, and average; the first preprint then took the greatest convex minorant of each curve before allocating. This is what one would write first, and what the first preprint did with $R=200$ \citep{bae2026luck}. The allocation is then optimized on the estimated curves and evaluated on the same curves (in sample).

\emph{Sample split.} Decide the allocation on curves built from one half of the samples and evaluate it on curves built from the other half. This removes the reuse of the same noise for selection and evaluation.

\emph{Exact curves.} The expected minimum of a uniformly random $k$-subset of the stored costs has a closed form,
\begin{equation}
\hat f(k) = \sum_{j=1}^{m-k+1} \frac{\binom{m-j}{k-1}}{\binom{m}{k}}\; x_{(j)}\,,
\label{eq:ustat}
\end{equation}
the U-statistic \citep{hoeffding1948ustat,david2003order} that the Monte Carlo average approximates, already standard in machine-learning evaluation \citep{tang2020showing} and, for binary outcomes, as the pass@$k$ estimator \citep{chen2021codex}. It is the Monte Carlo curve with $R\to\infty$, and it is unbiased for the population curve when the stored costs are independent draws. It removes the error of curve construction; it does not by itself remove the selection bias of optimizing and evaluating an allocation on the same samples, which Section~\ref{sec:whichest} measures.

\subsection{Which estimator is right}\label{sec:whichest}

\begin{figure}[t]
\centering
\includegraphics[width=0.75\textwidth]{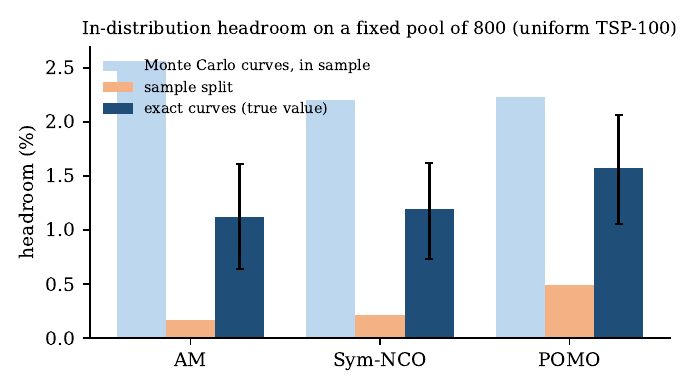}
\caption{Three estimators of the same quantity. Monte Carlo curves optimized in sample overstate the headroom; the sample split understates it; exact curves give the true value, with 95\% bootstrap intervals.}
\label{fig:estimators}
\end{figure}

One case allows a direct check. Under POMO-style decoding (greedy rollouts from every start node under eight augmentations) a 100-node instance has a deterministic pool of 800 trajectories, and the inference protocol draws from that fixed pool. The exact curve over the full pool is then the true curve of the protocol, and Eq.~\eqref{eq:headroom} on exact curves is the true headroom. Table~\ref{tab:estimators} and Fig.~\ref{fig:estimators} compare the three estimators on 50 uniform TSP-100 instances with $u=100$.

\begin{table}[t]
\caption{In-distribution headroom (\%) on a deterministic pool of 800 per instance (50 uniform TSP-100 instances, $u=100$). The exact value is the true headroom of the protocol; the interval is a 95\% instance bootstrap.}
\label{tab:estimators}
\begin{tabular}{@{}lccc@{}}
\toprule
Estimator & POMO & Attention Model & Sym-NCO\\
\midrule
Monte Carlo curves, in sample & 2.23 & 2.57 & 2.21\\
Sample split (mean of 20 random halvings) & 0.49 & 0.17 & 0.22\\
\textbf{Exact curves (true value)} & \textbf{1.57} [1.06, 2.07] & \textbf{1.12} [0.64, 1.61] & \textbf{1.19} [0.73, 1.62]\\
\botrule
\end{tabular}
\end{table}

The Monte Carlo estimator overstates the true headroom by 0.7--1.5 points. The source is the estimator, not the sampling of solutions: with 200 orderings the curves carry Monte Carlo noise, and the greedy allocation, which looks for the largest marginal reductions, finds that noise. This is the optimizer's curse \citep{smith2006curse,mak1999bounding}; the opposite bias of held-out evaluation has the same structure as in predictive optimization \citep{ito2018unbiased}. The sample split understates the headroom by a factor of three to seven: curves built from 400 samples are not the curves of the 800-sample pool, and evaluating on a different half penalizes allocations that are correct for the full pool. The first preprint's own split estimates (0.46, 0.02 and $-0.51$, with intervals covering zero) were read there as evidence that the in-distribution headroom is zero \citep{bae2026luck}. That reading is wrong; the true value is 1.1--1.6\%.

When samples are independent draws rather than a fixed pool (the stochastic sampling used in the rest of this paper), the exact curve is unbiased but the in-sample allocation still selects on sampling noise. We measured the size of this residual by simulation, treating each instance's 1,000 stored costs as the population, so that the true headroom is known (Appendix~\ref{app:sim}). With 800 samples per instance, $u=100$ and the 400-sample cap used throughout, exact-curve plug-in has a mean bias of $+0.16$ points and a root mean squared error of 0.54, against $-0.47$ and 1.00 for the sample split. On workloads in which every instance shares one population, so that the true headroom is zero, the plug-in reports $+0.15$ to $+0.52$ on average, with 95th percentiles up to 1.47. Section~\ref{sec:howmuch} uses full pools of 800 or 1,000 samples, for which the bias should be similar or slightly smaller. We use the exact plug-in throughout and treat values below about one point as indistinguishable from zero; none of the conclusions of Sections~\ref{sec:predict} and~\ref{sec:policy} depends on differences of that size.

\subsection{Keeping the axis and the outcome apart}\label{sec:split}

Section~\ref{sec:predict} relates the headroom of a workload to statistics of its instances, such as their difficulty. Both would be computed from the same samples if nothing were done, and an instance that looked difficult by sampling luck could look allocation-friendly by the same luck. Each instance therefore stores $K=1000$ samples, split once: samples 1--200 define its difficulty and any probe statistic, and samples 201--1000 define its exact curve and hence the headroom. No quantity on the horizontal axis of any analysis shares samples with the outcome. Sections~\ref{sec:predict} and~\ref{sec:policy} use this split, so their curves are built from 800 samples. Section~\ref{sec:howmuch} has no horizontal axis and uses each instance's full pool (800 deterministic or 1,000 stochastic samples).

\section{Experimental setup}\label{sec:setup}

\paragraph{Solvers.} Three pretrained TSP-100 checkpoints, POMO \citep{kwon2020pomo}, the Attention Model (AM) \citep{kool2019attention} and Sym-NCO \citep{kim2022symnco}, loaded through RL4CO \citep{berto2024rl4co}. Outside Section~\ref{sec:whichest}, solutions are stochastic decoding rollouts, the canonical inference mode of the Attention Model and the only one without an upper bound on the number of distinct solutions. Gaps are measured against LKH-3 tours \citep{helsgaun2017lkh3} computed and frozen before any allocation was evaluated. The per-instance budget is $u=100$ throughout, and no instance may receive more than 400 samples.

\paragraph{Difficulty, spread and level.} An instance's \emph{difficulty} (severity) is its mean gap over samples 1--200 divided by the mean of the same quantity over a reference group of 40 uniform 100-node instances. It is a multiple of the in-distribution gap: a severity of 10 means the solver's single-sample gap is ten times its typical in-distribution gap. For a workload, \emph{spread} is the standard deviation of log severity across its instances, and \emph{level} is the median severity.

\paragraph{Two heterogeneity mechanisms.} Workloads are made heterogeneous in two ways, both producing instances that a fixed solver finds unequally hard.
\begin{itemize}
\item \emph{Cluster knob.} 100 nodes around four Gaussian centres, with the cluster width $\sigma\sim\mathrm{logUniform}[0.03,0.35]$ drawn per instance. Tighter clusters are harder for all three solvers and have shorter optimal tours.
\item \emph{Size knob.} Uniform points with $n\in\{100,125,150,175,200\}$ drawn per instance, solved by the same 100-node checkpoint. Larger instances are harder and have longer optimal tours.
\end{itemize}
The two knobs were chosen because instance scale runs in opposite directions under them, which Section~\ref{sec:level} uses.

\paragraph{Collections.} Table~\ref{tab:collections} lists the data.

\begin{table}[t]
\caption{Collections. C1, C2A and C2B each contain 40 uniform reference instances and 240 heterogeneous instances; C2A and C2B are the two arms of the second confirmation, so C1 and C2 are the two confirmation collections. All store 1,000 stochastic samples per instance; the in-distribution collection also stores the 800-trajectory deterministic pool used in Section~\ref{sec:whichest}. Seeds are listed in Appendix~\ref{app:repro}.}
\label{tab:collections}
\begin{tabular}{@{}llll@{}}
\toprule
Name & Instances & Used in & Origin\\
\midrule
In-distribution & 50 uniform TSP-100 & Sect.~\ref{sec:whichest}, \ref{sec:howmuch} & \citep{bae2026luck}\\
Mixed & 50 uniform + 50 clustered & Sect.~\ref{sec:howmuch} & \citep{bae2026luck}, confirmation\\
C1 & cluster knob & Sect.~\ref{sec:predict} & \citep{bae2026spread}, confirmation 1\\
C2A & cluster knob & Sect.~\ref{sec:howmuch}, \ref{sec:predict} & \citep{bae2026spread}, confirmation 2\\
C2B & size knob & Sect.~\ref{sec:predict} & \citep{bae2026spread}, confirmation 2\\
Policy & 120 uniform + 120 clustered & Sect.~\ref{sec:policy} & \citep{bae2026spread}\\
\botrule
\end{tabular}
\end{table}

\paragraph{Workloads and statistics.} For Section~\ref{sec:predict}, each collection yields 200 workloads of 40 instances, each drawn from a random window of the severity-sorted pool, so that spread and level vary across workloads and are not perfectly correlated. Each workload is one observation of (spread, level, headroom). Associations are partial Spearman correlations computed by rank residualization. Confidence intervals are 95\% percentile intervals from a bootstrap over instances ($B=400$) \citep{efron1993bootstrap}: since workloads share instances, each replicate resamples instances and retains the workloads of which at least 60\% of members survive. Intervals for single headroom values bootstrap the instances of that workload.

\paragraph{Use of AI tools.} A large language model (Claude, Anthropic) was used to help write the re-analysis and figure scripts, to draft and edit the text, and to check numbers against the result files and references against their sources. All analyses were run on the stored data with the released scripts, and the author checked every result and every sentence.

\paragraph{Registered and exploratory analyses.} The workload construction, the statistics and the primary tests of Section~\ref{sec:predict} were fixed in pre-registrations before the collections were drawn. The registrations specified the sample-split estimator; the change to exact curves was made afterwards, for the reason given in Section~\ref{sec:whichest}, and changes the outcome variable only. Appendix~\ref{app:prereg} reports the registered verdicts under both estimators. Analyses not in a registration are labelled exploratory where they appear.

\section{How much does reallocation buy?}\label{sec:howmuch}

\paragraph{On the training distribution.} With the deterministic pool, the true headroom is 1.57\% for POMO, 1.12\% for the Attention Model and 1.19\% for Sym-NCO (Table~\ref{tab:estimators}). With stochastic sampling on the same 50 instances the exact plug-in gives 0.67\% [0.44, 0.90], 1.99\% [0.90, 2.99] and 2.49\% [1.74, 3.42]. These may include an upward bias of a few tenths of a point (Section~\ref{sec:whichest}), and POMO's value is within the range that the simulation produces when the true headroom is zero. Reallocation is not worthless on the training distribution, but the optimal allocation reduces the total gap by only 0.7--2.5\% relative to uniform.

\paragraph{On a mixed workload.} On 50 uniform and 50 clustered instances the headroom is 13.35\% [9.28, 19.00] for the Attention Model and 11.75\% [6.64, 17.03] for Sym-NCO. Both solvers degrade sharply on the clustered instances: their best-of-1000 gap there is 35 and 27 times their uniform one. POMO's rises only 2.4-fold, and its headroom is 0.50\% [0.36, 0.66].

\paragraph{In tour length.} Headroom is a share of the optimality gap (Section~\ref{sec:measure}), so its value in tour length depends on how large the gap is. Table~\ref{tab:lengths} translates it. On the training distribution the uniform gap is about 1\% for the Attention Model and Sym-NCO, and the optimal reallocation shortens the mean tour by 0.03\%: practically nothing. On the mixed workload the uniform gap is 16--26\%, and the optimal reallocation shortens the mean tour by 1.1--1.8\%; the deployable rule of Section~\ref{sec:policy} achieves 1.1--1.4\% on the policy collection. These are the savings available at no additional sampling cost, on workloads where the solver is out of its depth.

\begin{table}[t]
\caption{Headroom translated into mean tour length. Uniform gap: mean gap of the uniform allocation ($u=100$). Length: percentage reduction of the mean tour length. Length reduction is $100(\sum_i L_i(u)-\sum_i L_i(k_i))/\sum_i L_i(u)$ with $L_i(k)=\ell_i(1+f_i(k)/100)$ and $\ell_i$ the reference length, so short instances, whose gaps are large, carry less weight than in the headroom. POMO's large in-distribution gap reflects stochastic single-start sampling, which is far from its canonical multi-start greedy mode. Point estimates, exact curves.}
\label{tab:lengths}
\footnotesize
\begin{tabular}{@{}llccc@{}}
\toprule
Collection & Solver & Uniform gap (\%) & Headroom (\% of gap) & Length reduction (\%)\\
\midrule
\multirow{3}{*}{In-distribution} & AM & 1.40 & 1.99 & 0.03\\
 & Sym-NCO & 1.19 & 2.49 & 0.03\\
 & POMO & 9.19 & 0.67 & 0.06\\
\midrule
\multirow{3}{*}{Mixed} & AM & 26.38 & 13.35 & 1.84\\
 & Sym-NCO & 15.94 & 11.75 & 1.14\\
 & POMO & 15.11 & 0.50 & 0.01\\
\midrule
\multirow{2}{*}{Policy, oracle} & AM & 25.06 & 16.39 & 2.28\\
 & Sym-NCO & 17.93 & 15.69 & 1.60\\
\multirow{2}{*}{Policy, probe rule} & AM & 25.06 & 10.09 & 1.42\\
 & Sym-NCO & 17.93 & 10.40 & 1.07\\
\botrule
\end{tabular}
\end{table}

\begin{figure}[t]
\centering
\includegraphics[width=\textwidth]{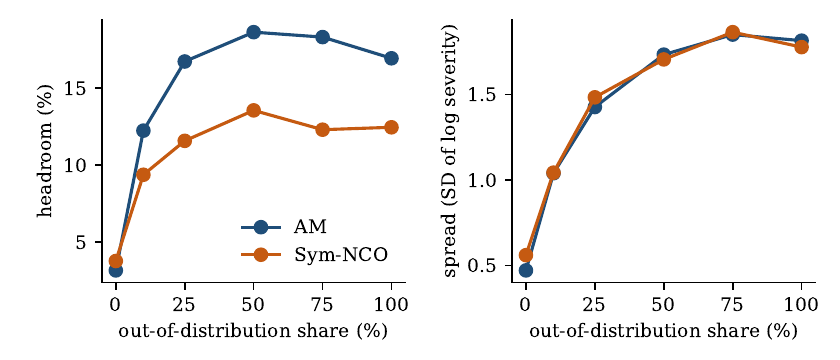}
\caption{Headroom (left) and spread of log severity (right) against the out-of-distribution share of 30-instance workloads (collection C2A, cluster knob; 25 workloads per share).}
\label{fig:composition}
\end{figure}

\paragraph{As a function of composition.} Fig.~\ref{fig:composition} varies the out-of-distribution share of 30-instance workloads drawn from C2A. For the Attention Model the headroom is 3.2\% with no clustered instances, 12.3\% at a 10\% share, 18.7\% at 50\% and 17.0\% when every instance is clustered; Sym-NCO rises similarly to a lower maximum (13.6\% at 50\%). The maximum is interior. The spread of difficulty (right panel) rises steeply with the first out-of-distribution instances and then levels off, while the headroom falls slightly at the highest shares, so the share and the spread do not carry the same information. Section~\ref{sec:predict} separates them.

\section{What predicts the gain: spread, not level}\label{sec:predict}

\subsection{Spread}\label{sec:spread}

\begin{figure}[t]
\centering
\includegraphics[width=0.62\textwidth]{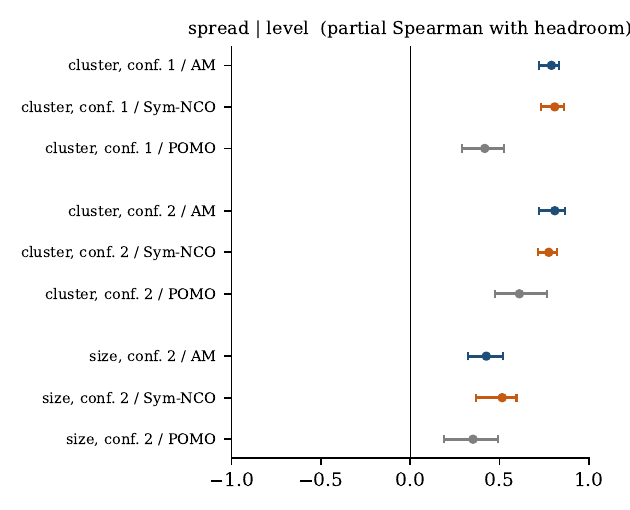}
\caption{Spread predicts the headroom in every cell: partial Spearman correlation of spread with headroom given level, with 95\% intervals.}
\label{fig:spread}
\end{figure}

Table~\ref{tab:partials} and Fig.~\ref{fig:spread} report, for each collection and solver, the partial rank correlation between a workload's spread and its headroom with level held fixed. All nine are positive and all nine intervals exclude zero. Under the cluster knob the correlation is 0.78--0.81 for the Attention Model and Sym-NCO in both confirmations. Under the size knob it is 0.43 [0.32, 0.52] and 0.51 [0.37, 0.59], smaller but clear; for the Attention Model and Sym-NCO the size knob produces spreads in 0.06--0.50 against 0.12--1.76 under the cluster knob, so a weaker association is expected.

The result does not depend on the budget. Repeating the C2A and C2B analyses at $u=25$, 50, 100 and 200 samples per instance (same workloads, same bootstrap), spread $\mid$ level is positive with an interval excluding zero in 23 of the 24 solver-knob-budget cells: 0.76--0.89 for the Attention Model and 0.63--0.90 for Sym-NCO under the cluster knob, and 0.32--0.50 and 0.42--0.61 under the size knob. The exception is POMO under the size knob at $u=200$, $+0.08$ [$-0.02$, $+0.20$], where its mean headroom is 0.3\%. The association weakens somewhat as the budget grows. For the Attention Model and Sym-NCO the level term described next is positive under the cluster knob and negative under the size knob at every budget (the interval for the Attention Model at $u=25$ includes zero).

The 200 windowed workloads share instances, and the instance bootstrap accounts for that dependence. As an additional check with independent observations, we cut the severity-sorted pool into 24 blocks of ten instances and paired the blocks at random into twelve disjoint workloads of twenty, so that spread varies across workloads while no instance is used twice; the rank-residualized partial correlation of spread with headroom given level is then computed over twelve independent points. Over 1,000 random pairings per cell, the median correlation is 0.65--0.75 for the Attention Model and Sym-NCO in all three collections, it is positive in at least 99\% of pairings, and a permutation test over the twelve workloads gives $p<0.05$ in 72--92\% of pairings despite having only twelve observations. For POMO the median is 0.51--0.73. These medians exceed the main partial correlations under the size knob (0.43--0.51) because pairing blocks from opposite ends of the severity order produces a wider range of spread than the windowed workloads.

\begin{table}[t]
\caption{Partial Spearman correlations of workload headroom with spread (given level), level (given spread) and median optimal tour length (given spread), with 95\% instance-bootstrap percentile intervals; because rank statistics are bounded and residualization is non-linear, point estimates need not lie at interval centres. 200 workloads of 40 instances per cell; exact-curve headroom.}
\label{tab:partials}
\footnotesize
\begin{tabular}{@{}llccc@{}}
\toprule
Collection & Solver & spread $\mid$ level & level $\mid$ spread & tour length $\mid$ spread\\
\midrule
\multirow{3}{*}{C1, cluster} & AM & $+0.79$ [$+0.72$, $+0.83$] & $+0.50$ [$+0.17$, $+0.71$] & $-0.49$ [$-0.67$, $-0.17$]\\
 & Sym-NCO & $+0.81$ [$+0.73$, $+0.86$] & $+0.76$ [$+0.48$, $+0.78$] & $-0.69$ [$-0.72$, $-0.45$]\\
 & POMO & $+0.42$ [$+0.29$, $+0.53$] & $+0.13$ [$-0.07$, $+0.29$] & $-0.13$ [$-0.28$, $+0.07$]\\
\midrule
\multirow{3}{*}{C2A, cluster} & AM & $+0.81$ [$+0.72$, $+0.87$] & $+0.58$ [$+0.29$, $+0.71$] & $-0.56$ [$-0.69$, $-0.28$]\\
 & Sym-NCO & $+0.78$ [$+0.71$, $+0.82$] & $+0.58$ [$+0.22$, $+0.63$] & $-0.56$ [$-0.61$, $-0.20$]\\
 & POMO & $+0.61$ [$+0.48$, $+0.76$] & $+0.31$ [$+0.13$, $+0.39$] & $-0.32$ [$-0.40$, $-0.15$]\\
\midrule
\multirow{3}{*}{C2B, size} & AM & $+0.43$ [$+0.32$, $+0.52$] & $-0.90$ [$-0.94$, $-0.81$] & $-0.90$ [$-0.94$, $-0.80$]\\
 & Sym-NCO & $+0.51$ [$+0.37$, $+0.59$] & $-0.88$ [$-0.90$, $-0.79$] & $-0.89$ [$-0.91$, $-0.81$]\\
 & POMO & $+0.35$ [$+0.19$, $+0.49$] & $-0.78$ [$-0.82$, $-0.65$] & $-0.78$ [$-0.83$, $-0.64$]\\
\botrule
\end{tabular}
\end{table}

POMO deserves a separate word. Its correlations are positive too, but its headroom is at most about 0.75\% in any workload of C1 or C2A and 0.40\% in C2B, against up to 22\% for the Attention Model. A correlation computed over a range of a few tenths of a percent says little about whether allocation is worth doing. POMO serves as a negative control in magnitude: the workload properties that make allocation valuable for the other two solvers do not make it valuable for a solver that the workload does not disturb.

\subsection{Level, and what it measures}\label{sec:level}

\begin{figure}[t]
\centering
\includegraphics[width=\textwidth]{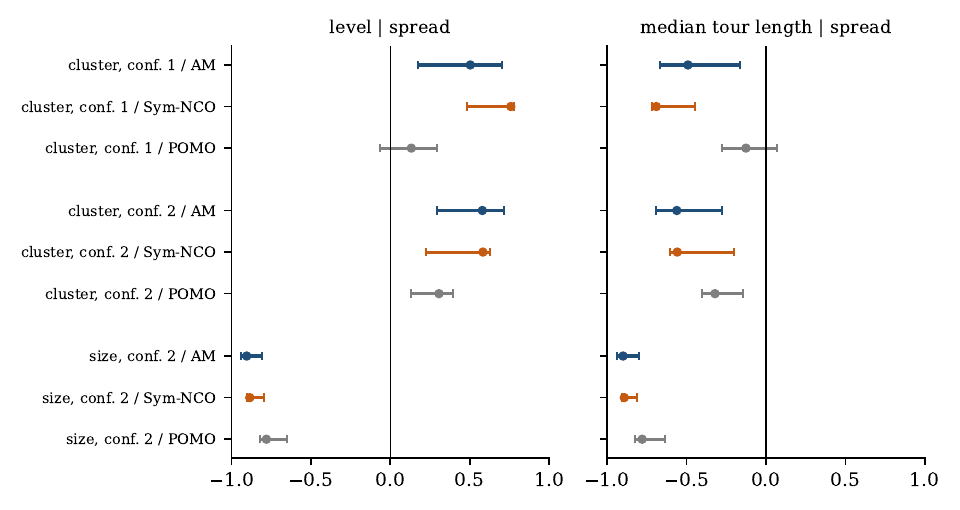}
\caption{The level term changes sign between the two knobs (left); the tour-length term does not (right). Partial Spearman correlations given spread, with 95\% intervals.}
\label{fig:level}
\end{figure}

If average difficulty mattered in its own right, the level term would have the same sign under both knobs. It does not (Table~\ref{tab:partials}, Fig.~\ref{fig:level}). Under the cluster knob, holding spread fixed, higher level goes with more headroom ($+0.50$ to $+0.76$ for the Attention Model and Sym-NCO). Under the size knob it goes with less ($-0.78$ to $-0.90$, all three solvers).

The two knobs differ in one known respect. Severity and instance scale are negatively related under the cluster knob (tight clusters are both harder and shorter) and positively under the size knob (large instances are both harder and longer). Replacing level by the workload's median optimal tour length gives a term that is negative in all nine cells and excludes zero in eight (Table~\ref{tab:partials}, last column; the exception is POMO in C1). The consistent reading is that workloads of shorter instances gain more from reallocation, once spread is fixed, and that the level term changes sign only because severity and scale trade places between the knobs.

Headroom is measured in gap units, percent of the optimal length, and a fixed absolute improvement is a larger gap on a shorter instance. The scale reading could therefore be partly a unit effect on the outcome side. As a check that was specified in the second preprint's registration before it was first run, but is not a registered endpoint, we recomputed the tour-length term with the outcome in absolute length units, $\frac1N\sum_i f_i(u)\,\ell_i/100 - \frac1N\sum_i f_i(k^\ast_i)\,\ell_i/100$, where $\ell_i$ is the reference length, which contains no length-bearing denominator. The term remains negative and excludes zero in three of four cells: $-0.90$ [$-0.91$, $-0.82$] and $-0.89$ [$-0.90$, $-0.74$] under the cluster knob for the Attention Model and Sym-NCO, and $-0.53$ [$-0.62$, $-0.30$] under the size knob for the Attention Model. For Sym-NCO under the size knob it is $-0.07$ [$-0.18$, $+0.09$]. The scale reading is therefore not a pure normalization artefact, but under the size knob it is not established for every solver. We present it as the most consistent interpretation of the level term, not as a finding.

\subsection{Composition adds little beyond spread}\label{sec:comp}

The share of out-of-distribution instances is the property one would most naturally monitor. With 30-instance workloads at shares of 0, 10, 25, 50, 75 and 100\% (25 workloads each, C2A), the partial correlation of share with headroom given spread is $+0.04$ [$-0.10$, $+0.13$] for the Attention Model and $+0.08$ [$-0.12$, $+0.20$] for Sym-NCO, while spread given share is $+0.43$ [$+0.20$, $+0.50$] and $+0.42$ [$+0.27$, $+0.51$]. Holding level fixed as well, share retains a small association for Sym-NCO, $+0.16$ [$+0.02$, $+0.25$], and none for the Attention Model, $+0.11$ [$-0.06$, $+0.18$]. The share is informative mainly through the spread it induces, and it is non-monotone (Fig.~\ref{fig:composition}).

\section{A deployable rule that pays for its own signal}\label{sec:policy}

The headroom of Section~\ref{sec:howmuch} assumes the curves are known. A deployable policy must learn about each instance from samples that count against the same budget. We evaluate single-pass rules of the following form on the policy collection (120 uniform and 120 clustered instances). Draw a probe of $m$ samples per instance; these are charged to the budget $S=uN$ and kept as candidate solutions. Split the remaining $S-mN$ samples in proportion to a statistic of the probe. Report the best of each instance's $k_i=m+\text{extra}_i$ samples. The probe statistics use raw tour costs only, so no reference tour is needed at deployment time. Probes are taken from samples 1--200 and scored on the exact curves of samples 201--1000. Allocations above the 400-sample cap are clipped and the excess is not redistributed, which slightly understates the probe rules (for Sym-NCO at $m=20$, 526 of 24,000 samples are lost).

\begin{table}[t]
\caption{Gain over uniform allocation (\%) on the policy collection, with 95\% instance-bootstrap intervals. Probe rules are charged ($m=20$); the oracle and the label rule are not. Recovery is the probe rule's gain as a share of the oracle.}
\label{tab:policy}
\footnotesize
\begin{tabular}{@{}lcccc@{}}
\toprule
 & Oracle & Label 20:1 & Probe CV & Probe $(\mathrm{med}-\min)/\mathrm{med}$\\
\midrule
AM & 16.39 [13.31, 19.84] & 9.02 [6.86, 11.28] & 4.91 [3.04, 7.22] & \textbf{10.09} [7.54, 12.89] (62\%)\\
Sym-NCO & 15.69 [13.64, 17.56] & 8.15 [6.64, 9.69] & 5.13 [3.29, 7.17] & \textbf{10.40} [8.61, 12.14] (66\%)\\
POMO & 0.51 [0.39, 0.61] & $-11.88$ [$-12.97$, $-10.25$] & $-0.20$ [$-0.30$, $-0.10$] & $-0.16$ [$-0.27$, $-0.02$]\\
\botrule
\end{tabular}
\end{table}

\begin{figure}[t]
\centering
\includegraphics[width=0.78\textwidth]{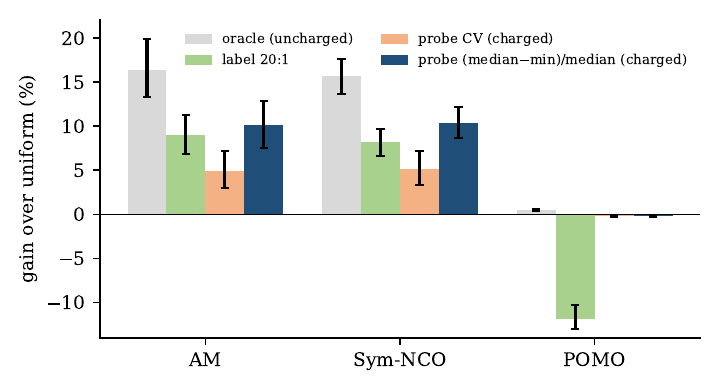}
\caption{Charged probe rules against the uncharged oracle and a distribution-label rule.}
\label{fig:policy}
\end{figure}

Table~\ref{tab:policy} and Fig.~\ref{fig:policy} compare two probe statistics: the coefficient of variation of the probe costs, used in the first preprint \citep{bae2026luck}, and the distance from the probe's median to its minimum relative to the median, which an exploratory attribution in the second preprint found to carry most of the signal \citep{bae2026spread}. That attribution used a model fitted on a different collection, but the statistic was chosen after seeing it, so the comparison is exploratory, as are the $m=40$ values and the paired comparison with the label rule. The second recovers 62\% of the oracle for the Attention Model and 66\% for Sym-NCO, and beats the coefficient of variation by $+5.18$ [$+3.97$, $+6.74$] and $+5.28$ [$+3.98$, $+6.54$] points in paired comparisons. With $m=40$ it gives 9.21\% and 9.51\%, so a smaller probe is slightly better at this budget. For reference, a rule that knows each instance's distribution label and gives clustered instances twenty times the samples of uniform ones, without paying for the label, obtains 9.02\% and 8.15\%. The probe rule is ahead of it by $+1.07$ [$-0.08$, $+2.42$] points for the Attention Model and $+2.25$ [$+1.15$, $+3.55$] for Sym-NCO, without needing the label.

The median-to-minimum statistic measures how much of the probe's mass sits above its best value, a crude indicator of how much a longer search could still find. The second preprint reported a learned policy over seven probe features with a gain of 10.46\% and 12.07\% \citep{bae2026spread}, measured with the earlier estimator and therefore not directly comparable. The single feature comes close for the Attention Model, falls short for Sym-NCO, and needs no training; we do not pursue the learned policy here.

On POMO every rule loses: the probe rules by about 0.2 points, the label rule by 11.9. Allocating by a label on a solver that the labelled distribution does not disturb starves half the instances for nothing, which is what convexity predicts when instances are close to exchangeable.

\paragraph{A sequential baseline.} A one-shot proportional split may look naive next to sequential allocation. We therefore also ran successive halving \citep{karnin2013sh}, adapted from arms to instances, under the same accounting: decisions use only samples 1--200, every sample is charged, and scoring uses the exact curves of samples 201--1000. Every instance first receives a floor of $f$ samples; half of the remaining budget is spent in $\lceil\log_2 N\rceil$ rounds, each giving the surviving instances equal shares and then dropping the half with the lowest score; the rest goes to the final survivors, with any excess over the 400-sample cap passed to the most recently dropped instances. With the median-to-minimum score and $f=20$, the best setting we tried, successive halving obtains 6.08\% [0.38, 12.45] for the Attention Model and 10.02\% [5.98, 13.23] for Sym-NCO, $-4.01$ [$-8.34$, $+1.01$] and $-0.39$ [$-3.27$, $+1.79$] points relative to the one-shot rule. Without a floor or with $f=50$ it does worse, and a score based on the improvement of the running minimum in the last round does much worse ($-19$ and $-21$ points of gain over uniform allocation). On POMO, successive halving with $f=20$ loses 7.63 points relative to uniform allocation, against 0.16 for the one-shot rule. Successive halving is designed to identify a few good alternatives and concentrates the budget on 30--60 of the 240 instances; when the objective is a sum over all instances, starving the rest costs more than the concentration gains. These comparisons are exploratory.

\paragraph{Model-based allocation.} A natural stronger competitor fits each instance's curve and allocates on the fit. Best-of-$k$ curves are well described by $f(k)=a+b\,k^{-\alpha}$ (Section~\ref{sec:discussion}), so we fit this form to the exact curve of each instance's probe ($k=1,\dots,20$, in cost relative to the probe minimum, so that no reference tour is needed), extrapolate it, and give the remaining budget greedily by the fitted marginal reductions, which is optimal for the fitted curves. A two-stage variant spends half of the remaining budget this way, refits each instance on all of its observed samples (at most 200, the decision slice), and allocates the other half on the refitted curves. Accounting and scoring are as above. The single-stage fit obtains 12.41\% [9.20, 16.11] for the Attention Model and 12.23\% [9.99, 14.89] for Sym-NCO; the two-stage fit obtains 14.36\% [11.16, 18.60] and 14.52\% [12.51, 16.48], 88\% and 93\% of the uncharged optimum and $+4.27$ [$+2.29$, $+6.19$] and $+4.12$ [$+2.70$, $+5.95$] points above the probe rule in paired comparisons. On POMO the two-stage fit loses 1.35 points against 0.16 for the probe rule: a curve model confidently extrapolates small differences that are not there. Model-based allocation is therefore the better rule when the workload is known to disturb the solver, and the probe rule is the safer one when it may not. This comparison was run after the review of an earlier version of this paper and is exploratory.

\section{Reallocation versus local search}\label{sec:ls}

Constructive neural solvers are often followed by a local search, and a solver that is 15--28\% from optimal on a workload would rarely be deployed without one. We therefore asked whether reallocation still matters once a cheap improvement step is applied. Because the stored arrays contain costs but not tours, we regenerated the mixed collection (the same 50 uniform and 50 clustered instances, from the original seed) and drew 400 fresh stochastic samples per instance with their tours, on CPU. With fresh samples on the regenerated instances, the headroom of reallocation on raw samples is 13.4\% for the Attention Model and 13.4\% for Sym-NCO, against 13.35\% and 11.75\% on the stored samples in Section~\ref{sec:howmuch}; the differences reflect new 400-sample pools in place of the stored 1,000-sample pools and lie within the intervals of Section~\ref{sec:howmuch}. The local search is 2-opt with best improvement to a local optimum, which takes about 3 milliseconds per tour.

\paragraph{Local search after the best sample.} Table~\ref{tab:ls} applies 2-opt to the best of each instance's samples under two allocations of the same budget, $u=100$: uniform, and the two-stage fitted-curve rule of Section~\ref{sec:policy}. Without local search, reallocation lowers the mean gap from 28.17\% to 24.30\% for the Attention Model and from 16.01\% to 14.44\% for Sym-NCO. Applying 2-opt to the uniform allocation's best tour lowers it to 2.53\% and 2.25\%, an order of magnitude more. After 2-opt, the reallocated allocation is slightly worse than the uniform one, by $+0.25$ [$+0.05$, $+0.45$] and $+0.19$ [$+0.04$, $+0.35$] points: the samples moved away from the uniform instances were worth more to 2-opt than the extra samples given to the clustered ones.

\paragraph{Local search on every sample.} A stronger pipeline applies 2-opt to every sample and keeps the best improved tour. We built exact best-of-$k$ curves on the improved costs and computed the oracle headroom as in Section~\ref{sec:howmuch}. The uniform allocation's mean gap falls to 1.02\% (Attention Model), 0.93\% (Sym-NCO) and 1.28\% (POMO). The oracle reallocation still removes a share of that gap, 6.3\% [4.4, 8.0], 6.6\% [4.9, 8.4] and 2.2\% [1.5, 2.8], but shortens the mean tour by only 0.039\%, 0.036\% and 0.039\%, against 1.88\% and 1.31\% without local search.

The practical reading is plain. Reallocation is worth doing only when the pipeline ends at the neural solver's samples. When any local search follows, the local search removes most of the gap at negligible cost, and what reallocation adds is a few hundredths of a percent of tour length. The findings of Sections~\ref{sec:howmuch}--\ref{sec:policy} describe the sampling stage itself; they do not argue for reallocation in a pipeline that includes local search.

\begin{table}[t]
\caption{Reallocation and 2-opt on the regenerated mixed collection ($u=100$). Mean gap (\%) of the best sampled tour, before and after 2-opt applied to it, with 95\% instance-bootstrap intervals; the last column is the paired difference after 2-opt (reallocated minus uniform).}
\label{tab:ls}
\footnotesize
\begin{tabular}{@{}lccccc@{}}
\toprule
Solver & Uniform & Uniform + 2-opt & Reallocated & Reallocated + 2-opt & Difference after 2-opt\\
\midrule
AM & 28.17 & 2.53 & 24.30 & 2.77 & $+0.25$ [$+0.05$, $+0.45$]\\
Sym-NCO & 16.01 & 2.25 & 14.44 & 2.44 & $+0.19$ [$+0.04$, $+0.35$]\\
POMO & 14.97 & 4.74 & 15.66 & 4.47 & $-0.27$ [$-0.57$, $+0.03$]\\
\botrule
\end{tabular}
\end{table}

\section{Discussion}\label{sec:discussion}

\paragraph{Recommendations.} For practitioners deploying a sampling-based neural routing solver under a fixed budget, the results suggest the following.
\begin{enumerate}
\item \emph{Add local search before reallocating.} On our workloads 2-opt on the best sample removes most of the gap in milliseconds, and reallocation adds nothing after it. Reallocate only when the samples are the final output.
\item \emph{Check whether the workload disturbs the solver.} If the solver's quality on the incoming instances is close to its in-distribution quality, reallocation reduces the total gap by at most about two percent; do not reallocate.
\item \emph{Judge by spread, not by average difficulty.} Holding spread fixed, the association with average difficulty changed sign between our two heterogeneity mechanisms, and may reflect instance scale.
\item \emph{Do not use the out-of-distribution share as the decision variable.} It works mainly through the spread it induces and is non-monotone.
\item \emph{Charge the probe, and fit the curves if the workload is disturbing.} A charged two-stage curve fit obtains about 90\% of the uncharged optimum on these workloads and a one-shot probe rule about two thirds; the fit loses more on an undisturbed solver. Report charged numbers: the uncharged optimum overstates what the simple rule can deploy by 50--60\%.
\item \emph{Measure with exact curves.} Monte Carlo curves optimized in sample overstate the gain, sample splitting understates it, and the exact U-statistic costs no more to compute than either.
\end{enumerate}

\paragraph{Continuous versus binary objectives.} In the language-model setting of the author's companion paper \citep{bae2026giveup}, each attempt either succeeds or not, and the value of one more attempt peaks at intermediate pass rates; there the average difficulty of a workload matters, because a workload of hopeless inputs offers nothing at any budget. Here the objective is the expected minimum of continuous costs, the marginal value of a sample never drops to zero, and what matters is how differently the instances' curves decay. The two results are consistent: which workload statistic governs reallocation depends on the shape of the objective, and transferring a rule of thumb from one setting to the other would be a mistake.

\paragraph{Curve shape.} In an exploratory fit to the full 1,000-sample pools of C2A and C2B, the exact curves are well described by $f(k)=a+b\,k^{-\alpha}$, with median $R^2$ of 0.996--0.999 per file. The severity spread used here is a proxy for differences in $b$ and $\alpha$ across instances. On those full-pool fits, the spread of $\log b$ predicts the headroom better than the severity spread under the size knob (rank correlation about $+0.86$ against $+0.48$ for the Attention Model), and the scale effect of Section~\ref{sec:level} suggests that $\alpha$ varies with instance size. We do not use $b$ here because estimating it from the 200-sample slice, as the separation of axis and outcome requires, was too noisy to improve on severity; a theory of the headroom in these parameters is a natural next step.

\section{Limitations}\label{sec:limits}

\begin{enumerate}
\item \emph{One problem class.} All results are for the TSP. The vehicle routing problem was excluded before collection because the reference solver's seed-to-seed variation exceeded our stability criterion. Nothing here establishes that the results transfer.
\item \emph{Observational design.} Workloads are constructed so that spread and level vary, but instances are not randomly assigned to difficulty. The associations are not causal effects.
\item \emph{The scale reading is post hoc} and holds in absolute units in three of four cells.
\item \emph{Residual bias.} Under stochastic sampling the exact plug-in headroom is biased upward by a few tenths of a point (0.15--0.52 on average when the truth is zero). This matters only for the in-distribution numbers, which it could account for in POMO's case.
\item \emph{The estimator change is post hoc relative to the registrations.} Appendix~\ref{app:prereg} reports both sets of verdicts.
\item \emph{Budget range.} The spread result holds for $u$ between 25 and 200 samples per instance (Section~\ref{sec:spread}), but the headroom levels, the composition sweep and the policy comparisons are reported at $u=100$ only.
\item \emph{Local search makes reallocation marginal.} With 2-opt the gain falls to a few hundredths of a percent of tour length. We tested only 2-opt; stronger local search would leave even less.
\item \emph{Older solvers, one size.} The three solvers are constructive models trained on 100-node instances; the size knob reaches 200 nodes. Larger and more recent solvers may degrade differently.
\end{enumerate}

\section{Conclusion}\label{sec:conclusion}

The uniform sampling budget used by neural routing solvers is close to optimal on the distribution they were trained on, and leaves 11.8--13.3\% of the gap on the table when the workload mixes in instances that degrade them. Whether a workload is of the second kind is predicted by the spread of difficulty within it, across two heterogeneity mechanisms, three solvers and two confirmation collections; its average difficulty is not a reliable guide. A charged probe of twenty samples per instance recovers about two thirds of the attainable gain with a single statistic, and about 90\% when a power-law curve is fitted to it. A cheap 2-opt step, however, removes most of the gap that reallocation addresses, and after it reallocation adds almost nothing, so these results describe the sampling stage rather than a case for reallocation in pipelines with local search. None of these numbers can be trusted unless the best-of-$k$ curves are computed exactly: the two estimators one would naturally use err in opposite directions, and one of them led the author's own earlier analysis to a wrong conclusion.

\backmatter

\bmhead{Acknowledgements} None.

\section*{Statements and Declarations}

\begin{itemize}
\item \textbf{Funding.} No funding was received for this research.
\item \textbf{Competing interests.} The author has no competing interests to declare.
\item \textbf{Data availability.} The stored cost arrays of all collections, the reference tour lengths and the pre-registration documents are available at \url{https://github.com/nepersoned/best-of-k-allocation}.
\item \textbf{Code availability.} The analysis code, including the exact-curve re-analysis and the figure scripts, is available at the same repository.
\item \textbf{Author contribution.} J.B. designed the study, collected the data, performed the analysis and wrote the manuscript.
\item \textbf{Use of generative AI.} The author used Claude (Anthropic), a large language model, to help write the re-analysis and figure scripts, to draft and edit the manuscript text, and to check numbers against the result files and references against their sources. All analyses were run on the stored data with the released scripts. The author reviewed every output and every sentence and takes full responsibility for the content.
\end{itemize}

\begin{appendices}

\section{Reproducibility}\label{app:repro}

\paragraph{Checkpoints.} \texttt{ML4TSPBench/\{POMO,AM,SYMNCO\}}, legacy-format RL4CO state dictionaries requiring key remapping (feed-forward \texttt{module.0/2}$\to$\texttt{module.lins.0/1}; decoder \texttt{logit\_attention}$\to$\texttt{pointer}; the AM archive's \texttt{baseline.*} copy excluded). Loading asserts zero missing and zero unexpected keys.

\paragraph{Seeds.} Instance/sampling seeds: in-distribution collection 20260812/11; mixed collection 20260813/77; C1 20260825/919; C2A 31415926/2718; C2B 27182818/1618; policy collection 20260813/415. Workload construction and bootstrap seeds are those of the registered analysis script.

\paragraph{Size knob.} Instance counts by size: 100: 58, 125: 52, 150: 48, 175: 37, 200: 45.

\paragraph{Compute.} Collecting the second preprint's data took under two hours on a single T4 GPU \citep{bae2026spread}. Every analysis runs on the stored arrays without a GPU; the full re-analysis takes a few minutes on a laptop.

\section{Simulation of estimator accuracy}\label{app:sim}

To measure the accuracy of the estimators under stochastic sampling, each instance's 1,000 stored costs in C2A and C2B were treated as the population, so that the true curve, being the expected minimum of $k$ independent draws from it, and hence the true headroom of any workload, is known in closed form. A simulated study draws $m=800$ costs per instance with replacement and estimates the headroom of a 40-instance workload with $u=100$ and a per-instance cap of 400, as in the main analyses, either in sample on exact curves of all 800 draws or by deciding on exact curves of one half and evaluating on the other.

\emph{Accuracy.} For each of the six files (two collections, three solvers) the workloads are the 40 reference instances and 24 windowed workloads of the severity-sorted pool, with eight studies each (1,200 studies in total). Pooled, the exact-curve plug-in has bias $+0.16$ and RMSE 0.54 points; the sample split has bias $-0.47$ and RMSE 1.00.

\emph{Null.} A workload of 40 copies of one reference instance's population has true headroom exactly zero. Over eight source instances and six studies each per file, the plug-in averages $+0.32$ and $+0.32$ (AM), $+0.30$ and $+0.52$ (Sym-NCO), and $+0.17$ and $+0.15$ (POMO) for C2A and C2B, with 95th percentiles of at most 1.47 (Sym-NCO, C2B). The script is \texttt{merged/estimator\_sim.py} in the repository.

\section{Pre-registered verdicts under both estimators}\label{app:prereg}

In every registration a primary endpoint was that the lower bound of the 95\% interval for spread $\mid$ level exceed zero for the Attention Model, with Sym-NCO as replication and POMO as negative control; the C2A registration added level $\mid$ spread as a second primary endpoint.

\begin{table}[!ht]
\caption{Registered spread endpoints under the registered estimator (sample split, Monte Carlo curves) and under exact curves.}
\label{tab:prereg}
\footnotesize
\begin{tabular}{@{}llcc@{}}
\toprule
Collection & Solver & Registered estimator & Exact curves\\
\midrule
C1, cluster & AM & $+0.744$ [$+0.672$, $+0.786$] pass & $+0.79$ [$+0.72$, $+0.83$] pass\\
C2A, cluster & AM & $+0.791$ [$+0.705$, $+0.845$] pass & $+0.81$ [$+0.72$, $+0.87$] pass\\
C2B, size & AM & $+0.157$ [$-0.030$, $+0.301$] \textbf{fail} & $+0.43$ [$+0.32$, $+0.52$] pass\\
C2B, size & Sym-NCO & $+0.233$ [$+0.110$, $+0.339$] pass & $+0.51$ [$+0.37$, $+0.59$] pass\\
\botrule
\end{tabular}
\end{table}

Under the registered estimator the size-knob endpoint failed for the Attention Model; under exact curves it passes. We report the failure as the registered verdict and the exact-curve result as a re-analysis. The registered negative-control condition for POMO (an interval containing zero) failed marginally in C2A under the registered estimator and, under exact curves, fails clearly in both C1 and C2A; Section~\ref{sec:spread} explains why we evaluate POMO by magnitude instead. The level term passed its registered endpoint in C2A under the registered estimator ($+0.544$ [$+0.308$, $+0.668$]) and passes under exact curves; Section~\ref{sec:level} explains why we do not interpret it as an effect of difficulty.

\section{Relation to the preprints}\label{app:preprints}

Table~\ref{tab:preprints} lists each claim of the two superseded preprints and its status after the re-analysis with exact curves.
\setcounter{table}{0}

\begin{table}[!ht]
\caption{Claims of the two preprints and their status here.}
\label{tab:preprints}
\footnotesize
\begin{tabular}{@{}p{0.47\textwidth}p{0.47\textwidth}@{}}
\toprule
Preprint claim & Status in this paper\\
\midrule
In-distribution headroom is not detectable; in-sample estimates of about 2\% are phantom \citep{bae2026luck} & \textbf{Corrected.} The true value is 1.1--1.6\%. The in-sample overstatement came from Monte Carlo curve noise, and the split estimate was biased downward.\\
Mixed workload gains 11--12\% \citep{bae2026luck} & Holds: 13.3\% and 11.8\% with exact curves.\\
Spread predicts the gain under the cluster knob \citep{bae2026spread} & Holds, slightly stronger.\\
Spread endpoint fails under the size knob for the Attention Model \citep{bae2026spread} & \textbf{Reversed}: passes with exact curves.\\
Level term reverses sign between knobs; scale reading holds in absolute units in 4 of 4 cells \citep{bae2026spread} & Reversal holds; scale reading holds in 3 of 4 cells.\\
Learned charged policy recovers 60--82\% \citep{bae2026spread} & Replaced by a single-feature rule recovering 62--66\% and a fitted-curve rule recovering 88--93\% with exact curves.\\
Ratio law for in-sample bias \citep{bae2026luck,bae2026spread} & Dropped; irrelevant once curves are exact.\\
\botrule
\end{tabular}
\end{table}

\end{appendices}

\bibliography{refs}

\end{document}